\documentclass[letterpaper]{article}
\usepackage{afterpage}
\usepackage{booktabs}
\usepackage{caption}
\usepackage{graphicx}
\usepackage{listings}
\usepackage{multicol}
\usepackage{tabularx}
\usepackage{url}
\usepackage{verbatim}
\usepackage[usenames]{color}
\usepackage{aaai}

\definecolor{ToDoColor}{rgb}{0.1,0.2,1}
\newcommand{\todo}[1]{\textbf{\color{ToDoColor} TODO: #1}}
\newcommand{\commentout}[1]{}

\begin{document}

\title{A Planning Approach to Monitoring Computer Programs' Behavior}
\author{Alexandre Cukier \\
Computer Science Dept.\\
Ben Gurion University\\
alexandre.cukier@gmail.com
\And 
Ronen I. Brafman\\
Computer Science Dept.\\
Ben Gurion University\\
brafman@cs.bgu.ac.il
\And
\ \ \ Yotam Perkal\\
\ \ \ PayPal, Israel\\
\ \ \ yperkal@paypal.com\\
\And 
David Tolpin\\
PayPal, Israel\\
dvd@offtopia.net \\
}

\maketitle
\begin{abstract}
    We describe a novel approach to monitoring high level behaviors using concepts from AI planning. Our goal is to understand what a program is doing based on its system call trace. This ability is particularly important for detecting malware. We approach this problem by building an abstract model of the operating system using the STRIPS planning language, casting system calls as planning operators. Given a system call trace, we simulate the corresponding operators on our model and by observing the properties of the state reached, we learn about the nature of the original program and its behavior. Thus, unlike most statistical detection methods that focus on syntactic features, our approach is semantic in nature. Therefore, it is more robust against obfuscation techniques used by malware that change the outward appearance of the trace but not its effect. We demonstrate the efficacy of our approach by evaluating it on actual system call traces.

\end{abstract}


\section{Introduction}

Malware is a serious threat for computer and Internet security for both  individuals and entities. 430 million new unique pieces of malware were detected by Symantec in 2015, and  94.1 millions of malware variants during only the month of February 2017.  Not surprisingly, to counter this threat, many techniques for malware detection have been proposed.
%
In this paper we are interested in the more general problem of understanding the behaviors taking place in the system. Given this information, one can determine whether they are malicious or not, and if malicious, provide an informed response. 

The standard approach to this problem is to use pattern-recognition methods, which are syntactic in nature. Roughly speaking, they view the input, whether code or events, as a long string of symbols, and seek properties of these strings that help classify them. To fool these methods, malware attempts to obfuscate its behavior by changing the sequence's properties~\cite{You:2010:MOT:1908631.1908903}.
Semantic methods, instead, try to model the underlying system, seeking to understand the input's meaning,
where in this paper, the input used is the system-call trace.%
\footnote{A system call is a mechanism used by a program to request from the operating system services it cannot perform directly, such access to hardware, files, network or memory.}
Therefore, they have the potential to be more robust to obfuscation attempts.

The most extreme and most accurate semantic approach is a faithful simulation of every trace followed by careful analysis of the resulting system state.
This is impractical: the analysis of the state of a computer following each trace is a non-trivial time consuming task that requires deducing high-level insights from the low level state and can only be conducted by experts.
%



Instead, we propose a methodology that uses an abstract system
model based on AI-planning languages and models. It requires a one-time, off-line effort by an expert, and can be used automatically to analyze each trace:
An expert that understands the semantics of system calls generates a planning operator for every system call. Each operator describes how the state of the system changes in response to the application of some system call.  Each operator 
is an abstraction that attempts to capture the system call's relevant effects. The abstraction process also involves the generation of a set of propositions describing the system state.  Now, given a system call trace, instead of simulating it
on the real system, we simulate the corresponding planning operators on the abstract state. The propositions true in the resulting state give us the needed information about what behaviors were carried out by this code.  This approach is fast and difficult to fool: obfuscation techniques that do not impact the actual behavior will not impact relevant aspects of the state.

In what follows we describe this methodology using examples, and
demonstrate its advantages by comparing it to statistical methods on 
actual system calls related to a mail application.



\section{Related work}

\cite{FHS96} is considered the seminal work which pushed
forward research on methods and representations of operating
system process monitoring based on system calls.
\cite{WFP99} provides an early comparison of machine learning
methods for modeling process behavior. \cite{GRS04} introduces
the model of execution graph, and behavior similarity-measure
based on the execution graph. \cite{MVV+06} combines multiple
models into an ensemble to improve anomaly detection.
\cite{XS10} applies continuous time Bayesian network (CTBN) to
system call processes to account for time-dependent features and
address high variability of system call streams over time.
\cite{KYL+16} applies a deep LSTM-based architecture to
sequences of individual system calls, treating system calls as a
language model.

Initially, only system call indices were used as
features~\cite{FHS96,WFP99}. \cite{LMH+05} compares three
different representations of system calls: $n$-grams of system call
names, histograms of system call names, and individual system
calls with associated parameters. \cite{PS07} proposes the use of
system call sequences of varying length as features.
\cite{LMH+05,TC06} investigate extracting features for machine
learning from arguments of system calls.  \cite{WSA+13} studies
novel techniques of anomaly detection and classification using
$n$-grams of system calls. \cite{CMK15} conducts a case study of
$n$-gram based feature selection for system-call based monitoring,
and analyses the influence of the size of the $n$-gram set and the
maximum $n$-gram length on detection accuracy.

Other work attempted to detect behaviors in a semantic way, using abstract representations of behaviors based on low level events and various techniques for detection. They all carry the notion of state, keeping track of effects of previous events. \cite{Christodorescu:2005:SMD:1058433.1059380} is the first to introduce semantics to characterize malicious behaviors. It builds behavior templates from binaries using formal semantics, which is used through a semantics-aware algorithm for detection. \cite{Martignoni:2008:LAD:1433006.1433013} builds multi-layered behavior graphs from low level events used through a behavior matcher. \cite{Jacob:2009:MBD:1691138.1691145} uses attribute grammars for abstraction and specification, using parallel automata for parsing and detection. \cite{5680136} specifies behaviors through UML activity diagrams from which one generates colored Petri Nets for detection. \cite{Beaucamps2012} uses first-order linear temporal logic to specify behaviors and model checking techniques for detection. \cite{Ezzati-Jivan:2012:SAG:2332427.2434003} offers an advanced state-full approach where behaviors are specified as finite state machines. 
Our approach is more fine-grained and general. We model the actual operators, not the target behaviors, although the model is informed by the behaviors. We illustrate this using the \textit{reverse shell} example in the next section.

Behavior recognition is closely related to plan and goal recognition~\cite{PlanRec}. Given a sequence of observed actions, the goal is to try to infer the actor's intentions. Typically, the output is a ranked list of hypothesized goals.
Most work assumes a library of possible behavior instances, i.e., plans, an approach limited in its ability to go beyond known instances. Probabilistic techniques, such as~\cite{Baker05} use Bayesian methods to assess the probability of various goals based on the actions involved. An influential recent approach is plan-recognition as planning~\cite{PRP1}, where the authors do away with the assumption of an explicit plan library. The plan library is replaced by a model of the domain (which implicitly defines the
set of possible plans), and the goal is to compute a good plan that is closest to the observed behavior. This line of work is appropriate when the observations are a subset of the actual actions taken, or when we attempt to recognize the goal before plan completion. We attempt to recognize malicious behavior off-line given a complete trace, although extensions for the online setting are natural.

\section{Background}

\subsection{AI Planning}

AI Planning is a decision making technique used  to find sequences of actions that  can transform a system from some initial state into a goal state.  
Formally, a classical planning problem is a tuple: $\pi=\langle P,A,I,G \rangle$. where:
$P$ is a  set of primitive propositions describing properties of interest of the system; $A$ is the action set.  Each action transforms the state of the system in some way; $I$ is the start state; and $G$ is the goal condition --- usually a conjunction of primitive propositions.
A state of the world, $s$, assigns truth values to all $p \in P$. Recall that a literal           is simply a primitive proposition or its negation.

An action $a\in A$ is a pair, \{\emph{pre}($a$), \emph{effects}($a$)\}, where 
\emph{pre}($a$) is a conjunction of literals, and
\emph{effects}($a$) is a set of pairs $(c,e)$ denoting conditional effects. 
We use $a(s)$ to denote the state that is obtained when $a$ is executed in state $s$.
If $s$ does not satisfy all literals in \emph{pre}($a$), then $a(s)$ is undefined.
Otherwise, $a(s)$ assigns to each proposition $p$ the same value as $s$, unless
there exists a pair $(c,e)\in $ \emph{effects}($a$) such that $s\models c$ and
$e$ assigns $p$ a different value than $s$.
We assume that $a$ is well defined, that is, if
$(c,e)\in\mbox{\emph{effects}}(a)$ then $c \wedge $\emph{pre}($a$) is
consistent, and that if both $(c,e),(c',e')\in $ \emph{effects}($a$) and
$s\models c\wedge c'$ for some state $s$, then $e\wedge e'$ is consistent.

The classical planning problem is defined as follows: given a planning problem $\pi$,
find a sequence of actions $\{a_1,\ldots,a_k\}$ (a.k.a.~a plan) such that
$a_k(\cdots(a_1(I))\cdots)\models G$.

To illustrate this model, consider a simplified domain with three action types: {\em socket}, {\em listen}, and {\em accept}. These actions model the effect of system calls that create a socket, listen for an incoming connection, and accept a connection. For the sake of this example, we ignore various parameters of these system calls, and assume that system calls do not fail.


The set $P$ contains: \{\textit{(opened socket-descriptor), (listening socket-descriptor), (connected socket-descriptor)}\}, where {\em socket-descriptor} is a parameter that we abbreviate as {\em sd}. 
The set of actions is:
\begin{itemize}
\item {\em socket(returned-sd)} with precondition: {\em $\neg$(opened returned-sd)}, and the effect: {\em (opened returned-sd)}%
\footnote{A more faithful model will use conditional effects instead, and will also consider their return value.}
\item {\em listen(sd)} has no precondition and the conditional effect: {\em (listening sd)} when {\em (opened sd)} $\wedge\neg${\em (listening sd)}

\item {\em accept(sd, returned-sd)} has the preconditions: {\em (listening sd)} and the effects: {\em(opened returned-sd)} and {\em(connected returned-sd)}
\end{itemize}
The plan {\em socket(sd1), listen(sd1), accept(sd1, sd2)} is a legal plan. Initially, all propositions are false. Because {\em (opened sd1)} is false,
we can apply {\em socket(sd1)}.
Once applied, {\em (opened sd1)}, the precondition of {\em listen},
becomes true. This results in {\em (listening sd1)} becoming true. Finally, {\em accept} needs a socket descriptor in the state {\em listening} ({\em sd1}) and another having  {\em opened} false
({\em sd2}). 
It now sets {\em sd2} to {\em opened} and {\em connected}. Given the resulting state, we recognize that a host connected itself to our local server.


On the other hand, the plan: {\em socket(sd1), accept(sd1, sd2)} is invalid because the preconditions of accept are not all sastified: {\em (listening sd2)} is not set to true.

Typically, planning models are used for generating plans. Thus,
in the above example, a planning algorithm could find the (abstracted) sequence of system calls required to achieve various goals. Our focus in this paper is on the planning model itself --- the propositions and the operators, as an abstraction of the operating system. The acting agent is a running process, the OS is the environment in which it is acting, and its system-call trace defines the plan, via our mapping. To determine what the process is doing, we simply observe the abstract state of the OS. For the purpose of this paper, we consider that the OS abstraction has a unique running and single thread process.

\section{Our approach}

We propose to build an abstract system model and simulate 
an abstraction of the system call trace on it.

The manual part of our approach is the construction of the abstraction. 
We associate an action with each system call, with preconditions (typically empty) and effects (typically conditional).  The set of propositions that we use to describe the system is informed by the type of behaviors we want to capture.  For example, whether channels were opened, files accessed, information transmitted over a channel, etc.
An action describes what new facts will become true following the execution of the system call it models, possibly conditional on other facts being true prior to its execution.

We illustrate this using the example of a remote shell: a command line interface controlled by a remote host often used by attackers to execute system commands. We focus on the \textit{reverse shell}, where a host connects itself to a remote server.
Starting a reverse shell requires a few steps: (1) Create a socket.
(2) Independently connect the socket to an endpoint and duplicate the socket descriptor to the standard input and output (so that the input and output streams  go through the socket).
(3) Execute a shell.

We use system calls {\em socket}, {\em connect}, {\em dup}, {\em fcntl}, {\em close} and {\em execve} that, respectively, create a socket, connect a socket to a remote host, duplicate a socket, set properties to a socket, close a socket, and execute a program. 
\vspace{1mm}
\noindent{\bf Propositions.} The propositions are: {\em(opened fd)}, {\em(is-socket fd)}, {\em(equal-fds fd1 fd2)}, {\em(close-on-exec fd)}, {\em(connected sd)}, {\em(is-shell path)}, {\em(remote-shell-started)}

\vspace{1mm}
\noindent{\bf Initial state.} The initial state initiates the resources used by a process when it starts, and taints the ones that have targeted properties:
\begin{itemize}
\item Propositions {\em (opened fd0)}, {\em (opened fd1)}, {\em (opened fd2)} are set to true, as fd0/1/2 denote standard input/output/error,
respectively, and these files are open.
\item Shell executable paths are tainted. We assume that we know all of those presents on the operating system. We handle two of them in this example: {\em /bin/sh} and {\em /bin/bash} that we name respectively {\em sh} and {\em bash}. Thus,  {\em (is-shell sh)} and {\em (is-shell bash)} are set to true.
\end{itemize}

\noindent{\bf Actions.} Planning operators are a simplified abstraction of the system calls. Since system calls called with wrong arguments do not make programs crash, and have no effect, the corresponding actions use conditional effects only -- i.e., they are always executable but change the state only if their conditions are met. 
\begin{itemize}
\item {\em socket(returned-sd, cloexec)} has the effects:
\\The flag {\em FD\_CLOEXEC} is represented by the boolean {\em cloexec}. 
\begin{itemize}
\item {\em(opened returned-sd)}$\wedge${\em(is-socket returned-sd)} if $\neg${\em (opened returned-sd)}
\item {\em(close-on-exec returned-sd)} if $\neg${\em (opened returned-sd)}$\wedge${\em (= cloexec True)}  
\end{itemize}
\item {\em connect(sd)} has the effects:
\begin{itemize}
\item {\em(connected sd)} if {\em (opened sd)}$\wedge${\em(is-socket sd)}$\wedge\neg${\em(connected sd)}
\item $\forall fd,$ {\em(connected fd)} if {\em(equal-fds sd fd)}$\wedge${\em (opened sd)}$\wedge${\em(is-socket sd)}$\wedge\neg${\em(connected sd)}
\end{itemize}
\item {\em dup(sd, returned-sd)} has the effects:
\begin{itemize}
\item {\em(opened returned-sd)}$\wedge${\em(equal-fds sd returned-sd)}$\wedge${\em(equal-fds returned-sd sd)} if {\em (opened sd)}$\wedge\neg${\em (opened returned-sd)}
\item {\em(is-socket returned-sd)} if {\em (is-socket sd)}$\wedge${\em (opened sd)}$\wedge\neg${\em (opened returned-sd)} 
\item {\em(connected returned-sd)} if {\em (connected sd)}$\wedge${\em (opened sd)}$\wedge\neg${\em (opened returned-sd)}
\item $\forall fd,$ {\em(equal-fds fd returned-sd)}$\wedge${\em(equal-fds returned-sd fd)} if {\em(equal-fds fd sd)}$\wedge\neg${\em (opened returned-sd)}
\end{itemize}
\item {\em fcntl(sd, command, returned-sd, cloexec)} has the effects: 
\\{\em returned-sd} is the argument of the command {\em F\_DUPFD} and {\em cloexec} is the argument of the command {\em F\_SETFD}. {\em F\_DUPFD\_CLOEXEC} uses both.
\\The flag {\em FD\_CLOEXEC} is represented by the boolean {\em cloexec}.
\begin{itemize}
\item same effects as {\em dup(sd, returned-sd)} if {\em (= command F\_DUPFD)}$\vee${\em (= command F\_DUPFD\_CLOEXEC)}
\item {\em(close-on-exec sd)} if [[{\em (= command F\_SETFD)}$\wedge${\em (= cloexec True)}]$\vee${\em (= command F\_DUPFD\_CLOEXEC)}]$\wedge${\em (opened sd)}$\wedge\neg${\em (opened returned-sd)}
\item $\neg${\em(close-on-exec sd)} if {\em (= command F\_SETFD)}$\wedge${\em (= cloexec False)}$\wedge${\em (opened sd)}$\wedge\neg${\em (opened returned-sd)}
\end{itemize}
\item {\em close(sd)} has the effects: 
\begin{itemize}
\item $\neg${\em (opened sd)}$\wedge\neg${\em (is-socket sd)}$\wedge $ $\neg${\em (connected sd)}$\wedge\neg${\em (close-on-exec sd)}
\item $\forall fd,$ $\neg${\em(equal-fds sd fd)}$\wedge\neg${\em(equal-fds fd sd)}
\end{itemize}
\item {\em execve(path)} has the effect: 
\begin{itemize}
\item {\em (remote-shell-started)} if {\em (is-shell path)} $\wedge\exists fd,$  {\em (connected fd)}$\wedge\neg${\em (close-on-exec fd)}$\wedge$[{\em (= fd fd0)}$\vee${\em (equal-fds fd fd0)}]$\wedge$[{\em (= fd fd1)}$\vee${\em (equal-fds fd fd1)}]
\end{itemize}
\end{itemize}

\subsubsection{Valid plans}

\begin{figure}
\hspace{-0.6cm}
      \includegraphics[scale=0.36]{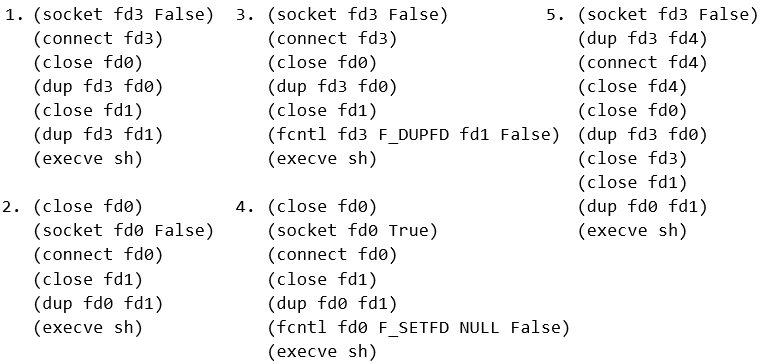}
    \caption{Valid Plans for the Reverse Shell domain}
        \label{fig:plans}
    \vspace{-0.5cm}
\end{figure}
\commentout{
\begin{enumerate}
\item \begin{verbatim}(socket fd3 False)
(connect fd3)
(close fd0)
(dup fd3 fd0)
(close fd1)
(dup fd3 fd1)
(execve sh)
\end{verbatim}
\item \begin{verbatim}(close fd0)
(socket fd0 False)
(connect fd0)
(close fd1)
(dup fd0 fd1)
(execve sh)
\end{verbatim}
\item \begin{verbatim}(socket fd3 False)
(connect fd3)
(close fd0)
(dup fd3 fd0)
(close fd1)
(fcntl fd3 F_DUPFD fd1 False)
(execve sh)
\end{verbatim}
\item \begin{verbatim}(socket fd3 False)
(dup fd3 fd4)
(connect fd4)
(close fd4)
(close fd0)
(dup fd3 fd0)
(close fd3)
(close fd1)
(dup fd0 fd1)
(execve sh)
\end{verbatim}
\item \begin{verbatim}(close fd0)
(socket fd0 True)
(connect fd0)
(close fd1)
(dup fd0 fd1)
(fcntl fd0 F_SETFD NULL False)
(execve sh)


\end{verbatim}
\end {enumerate}
}

The five different valid plans shown in Figure~\ref{fig:plans} show how diverse the plans are even for such a simple example.
Plan 1 is the standard sequence performed to establish a reverse shell, which appears in most shellcode databases.  Plan 2 uses the fact that we know that  system call {\em socket} allocates the lowest file descriptor available. Calling {\em close(fd0)} before {\em socket} avoids the duplication of the socket on the file descriptor 0. Plan 3 replaces one system call by an equivalent one:
{\em dup} is replaced by {\em fcntl} called with the command {\em F\_DUPFD}. Plan 4 demonstrates that planning captures and updates correctly properties set by flags and through different system calls. The flag {\em FD\_CLOEXEC} is first set through system call {\em socket}, and reset later by {\em fcntl} called on {\em F\_SETFD}.
Plan 5 shows that planning is able to follow complex flow of operation on file descriptors. The key point is that, despite major differences in appearance, which are likely to fool syntactic methods (certainly, if some of the plans were not available previously), our semantic approach recognizes the behavior they implement.

The main effort required by our approach is building an appropriate model for each system call. This  model is informed by the basic set of low-level behaviors one would like to model.  Once completed, we can simulate any sequence of system calls by applying them to an initial state of the abstract system using any planning simulator/validator. By examining the final state of the system, we can recognize which behaviors took place.  Thus, the off-line modeling task is done once, and the resulting model can be used repeatedly, automatically, and very cheaply,  to analyze programs. 

The (manual) abstraction process is flexible.  We can use it to identify simple behaviors, such as create a socket, connect to a remote host, read data from socket, open file for writing, write into file, etc.
And we can also recognize complex behaviors by detecting combinations of simple behaviors. For example, downloading a file requires reading data from a connected socket and writing it to an opened file. Thus, once we have the low-level behaviors, it is easy to capture the higher level ones. We can do this by either modifying the action model or by adding axioms, which are a method of adding a simple form of inference to planning.
With such a layered approach, basic behaviors can be reused to identify multiple high level behaviors.

As this model is an abstraction, some information is lost in this model,
and the method cannot be 100\% correct and capture every nuance.
Much can be captured by building a more elaborate model, but some aspects, such as accurate modeling of system resources, are not likely to be practical.

\section{Empirical evaluation}

In the previous section we demonstrated the capabilities of our approach to recognize behaviors on the reverse shell domain, where our planning model is able to recognize the same behavior generated in different ways.
We now want to
highlight our ability to recognize complex, higher level behaviors that are built from lower level behaviors, compared to statistical methods that are quite popular in this area. To do this, we consider the behavior of real processes involved in a mail service.  Given the system call trace logs of several processes, we attempt to recognize which behavior is realized by each of the processes, such as sending an email via SMTP, collecting an email from a remote server via IMAP, and so on. The code and data set used for the empirical evaluation can be obtained from a Git repository at \url{https://github.com/alexEnsimag/planning-for-syscall-monitoring}.

\subsection{Data Collection}

\begin{figure}
    \centering
      \includegraphics[scale=0.28]{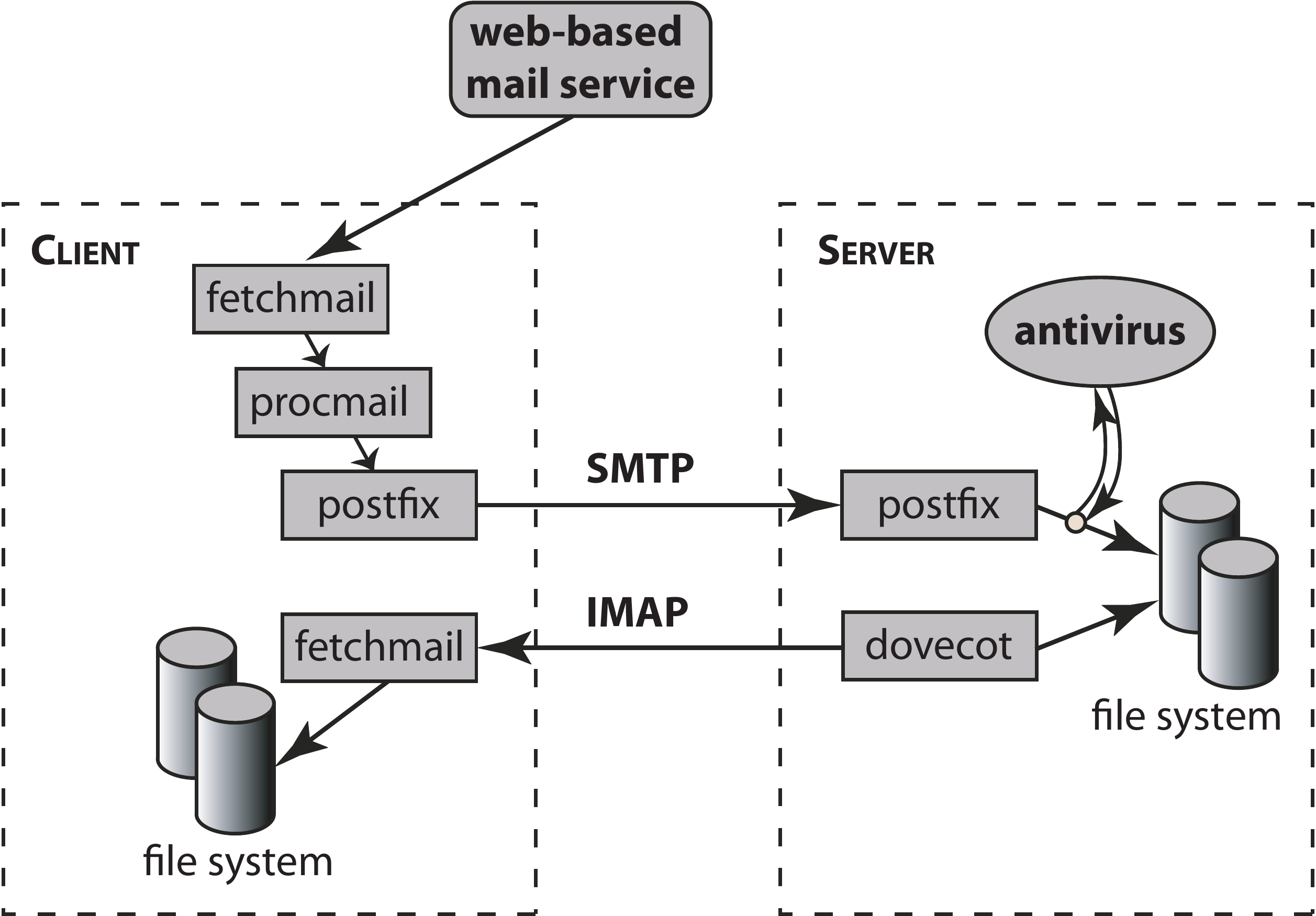}
    \caption{A mail service setup for evaluation of planning approach on mail delivery activities}
    \label{fig:mailservice}
    \vspace{-5mm}
\end{figure}

We generate system call traces of processes running in a mail
service (Figure~\ref{fig:mailservice}).  The setup consists of
two hosts: the client and the server, and involves a number of processes,
denoted in what follows in italic.
The hosts collect emails from an external
server. In order to provide sufficient volume and diversity of
the data processed, we opened a dedicated email account with a
web-based email service, and subscribed to
multiple promotion and notification mailing lists. On the
client, \textit{fetchmail} is used to retrieve emails from a
web-based email provider via the IMAP protocol.  Then,
\textit{procmail} dispatches received emails, which are then
sent by \textit{postfix} to the server via SMTP protocol.  The
server's \textit{postfix} process receives the emails, passes
them through the \textit{amavis} antivirus and stores in the
local filesystem. The \textit{dovecot} process serves emails via
the IMAP protocol. The emails are retrieved by the client's
\textit{fetchmail}, and stored in the filesystem.  We use
\textit{Docker}~\cite{Docker}  to run containers encapsulating
the mail server and mail client hosts, and
\textit{sysdig}~\cite{sysdig} to record the system calls.

We analyze system call traces of the following processes:
\textit{smtpd}, \textit{fetchmail} on the client, 
\textit{fetchmail} on the server,
\textit{imap-login},
all other processes.

These processes realize the following behaviors:
\begin{itemize}
    \item receiving an email over the SMTP protocol;
    \item receiving an email over IMAP protocol;
    \item forwarding an email from the client to the server;
    \item IMAP connection setup and authentication;
    \item other behaviours not  tracked by the system.
 \end{itemize}
We use 440 samples for each process (including \textit{other}). Data is split into training and test sets as 66\%/33\%.\\


\noindent{\bf Statistical Classification.} \ 
We compare our goal-tracking approach to a baseline, commonly used statistical  classification of processes based on system call sequences. We train a statistical classifier (random forest) on the collected system call traces. This classification approach is similar to approaches used in system call monitoring literature~\cite{Canali:2012:QSA:2338965.2336768}, ~\cite{Firdausi:2010:AML:1917825.1918195}. We use bi-gram vectorization of the system call names. Our vocabulary contains 100 most recurrent bi-grams in the corpus. Thus, each sample is represented by a vector of 100 elements, where each element in the vector represents one of the bi-grams in the vocabulary and the values represent the \textit{number of times} a specific bi-gram appeared in that sample. For example, if the first 4 elements in the vector correspond to
bi-grams \texttt{(open, read)}, \texttt{(read, write)}, \texttt{(write, read)}, and \texttt{(write, close)}, a system call trace
\begin{verbatim}
open, read, write, read, write, close
\end{verbatim}
will produce a bi-gram vector: $1, 2, 1, 1,\ldots$

We proceed with the empirical evaluation as follows:

\begin{enumerate}
\item We train a classifier that classifies each of the behaviors based on non-obfuscated system call sequences. The classifier achieves 97\% accuracy.
\item We create obfuscated samples in a way that `breaks' the bi-grams by inserting a  system call that has no effect on the process behavior (for example, \texttt{sleep} with a sufficiently small argument) in between each couple of system calls in the sequence. This method is called {\em adding semantic no-ops} and is the focus of~\cite{RosenbergG17}.
When testing the statistical model on the obfuscated data we  get 0\% accuracy (all  samples  in the test set are classified as `other').
\item We retrain our model on both obfuscated and non-obfuscated data. It now achieves $\approx 66\%$ accuracy.
\end{enumerate}

\noindent{\bf Planning-based Classification} \ 
Information contained in the system logs and manual inspection of system call traces in the training set are used to specify the planning domain and the goal for each of the behaviors. Then, the VAL plan validation system~\cite{VAL} is used to classify system call traces in the test set. The planning-based classifier based on the domain built for the original, non-obfuscated system call logs is applied to both non-obfuscated and obfuscated system call traces. In both cases, the planning based classifier reaches \textbf{over 98\% accuracy}.
Manual inspection of the misclassified samples suggests that the samples correspond to failed communication between components of the mail service.


\begin{table}[h]
\centering
\begin{tabular}{l|r|r}
              & Random forest & Planning-based \\ \hline
    Original & 97\% & \textbf{98\%} \\
    Obfuscated & 0\% & \textbf{98\%} \\
    Re-trained & 66\% & \textbf{98\%} 
\end{tabular}
\caption{Classification accuracy on real system logs}
\label{tbl:results}
\vspace{-2mm}
\end{table}

Table~\ref{tbl:results} summarizes the results.
Our goal-tracking approach successfully classifies processes.
Statistical methods are able to distinguish between different
processes, but are prone to obfuscation. The obfuscation
technique we used in this case study is particularly challenging
for statistical classification, and with other obfuscation techniques
the difference in accuracy might not be as sharp.  However, 
the planning-based approach is inherently more robust in the 
face of obfuscation, as it captures semantic behavior,
which must remain intact, rather than just statistical 
manifestations, which can be easily altered.

\section{Discussion and future work}
We presented an approach for monitoring computer programs using an abstract model of the system state and the basic "actions" that operate on this state --- system calls, in our case. The method is semantic in nature, and hence not prone to the weaknesses of syntactic methods that consider the command sequences form rather than their meaning. Unlike statistical methods that, in principle, can be fully automated, our approach has a non-trivial, one-time manual modeling step.
But once the model is constructed, it can be used automatically and with little cost. 

We demonstrated the effectiveness of our method by first showing how we capture a simple low level behavior that has diverse implementations using a simple model. Syntactically, each implementation is quite different, yet the common semantics can be captured by modeling just a few system calls. Then, we showed how we detect  more complex, higher level behavior with almost perfect accuracy, without being affected by obfuscation techniques that easily fool state-of-the-art statistical methods. 
%
%
The approach used here can be used for other applications beyond system-call logs, such as analysis of transactions,  HTTP logs, and more.
Moreover, we believe that it could complement statistical methods by allowing us to run statistical analysis on the higher level features generated by our abstract state.

\bibliographystyle{aaai}
\bibliography{bibli}

\end{document}